%% file: main.tex
\documentclass[sigconf, nonacm]{acmart}

\usepackage{pvldb}

\usepackage[ruled,vlined,linesnumbered]{algorithm2e}
\usepackage{placeins}
\usepackage{colortbl}
\usepackage{tabularx}
\usepackage{pifont}
\usepackage{hyperref}

\newcommand{\nlplsql}{NL-to-PL/SQL}
\newcommand{\nlsql}{Text-to-SQL}
\newcommand{\benchmark}{ProcArena}

\newcommand{\dmode}{Direct}
\newcommand{\imode}{Interactive}

\newcommand{\solver}{Solver}
\newcommand{\usersimulator}{User Simulator}
\newcommand{\ssprotocol}{\solver{}--\hspace{0pt}\usersimulator{}}



\let\modeband\tabband

\definecolor{artifactblue}{HTML}{1A5A8A}
\newcommand{\artifact}[1]{\textcolor{artifactblue}{\emph{#1}}}

\newcommand{\zebra}{\rowcolor{black!3}}

\newcolumntype{R}[1]{>{\raggedright\arraybackslash\hsize=#1\hsize}X}

\renewcommand\vldbdoi{XX.XX/XXX.XX}
\renewcommand\vldbpages{XXX-XXX}
\renewcommand\vldbavailabilityurl{https://github.com/ZhanGHanG9991/ProcArena} 

\begin{document}
\title{ProcArena: A Multi-Scenario Benchmark for LLMs on Direct and Interactive PL/SQL Development from Natural Language [Experiment, Analysis \& Benchmark]}

\author{Hang Zhang$^{1}$, Chaokun Wang$^{1, *}$, Yuzhi Pan$^{1}$, Ziyao Zhong$^{1}$, Shuo Cao$^{1}$,\\Yue Xue$^{1}$, Zeyu Huang$^{1}$, Xingwei Zhou$^{1}$, Fang Niu$^{1}$, Bofan Xie$^{1}$,\\Guanchen Ge$^{1}$, Leqi Zheng$^{1}$, Ziyang Liu$^{1}$, Xiannian Cao$^{2}$, Pengcheng Ge$^{2}$}

\affiliation{
    \institution{$^1$Tsinghua University, Beijing, China \quad $^2$Lenovo Group Limited, Beijing, China}
    \country{}
}

\email{\{zhanghang24, panyz24, zhongzy25, caos24, y-xue24, huangzy24, zhouxw24, nf21, xbf25, ggc22,\\zhenglq24, liu-zy21\}@mails.tsinghua.edu.cn, chaokun@tsinghua.edu.cn, \{caoxn3, gepc1\}@lenovo.com}

\renewcommand{\authors}{Hang Zhang, Chaokun Wang, Yuzhi Pan, Ziyao Zhong, Shuo Cao, Yue Xue, Zeyu Huang, Xingwei Zhou, Fang Niu, Bofan Xie, Guanchen Ge, Leqi Zheng, Ziyang Liu, Xiannian Cao, Pengcheng Ge}







\begin{abstract}
Large language models (LLMs) have shown strong potential for translating natural-language (NL) requirements into PL/SQL programs, attracting increasing attention from the database community. However, existing NL-to-PL/SQL efforts primarily focus on directly generating PL/SQL from complete NL requirements. In practice, PL/SQL development involves diverse scenarios, such as from-scratch development, code modification, debugging, and optimization, and may require either direct generation or multi-turn interaction. Yet, no comprehensive benchmark evaluates multi-scenario, direct and interactive, and multi-dialect NL-to-PL/SQL development. In this paper, we present \benchmark{}, an execution-based benchmark covering both \dmode{} and \imode{} modes. \benchmark{} comprises 3{,}998 executable tasks over 157 databases, spanning nine development subscenarios in PostgreSQL and Oracle. We construct challenging \dmode{} tasks through Iterative Logic Enhancement and scenario-specific adapters, and derive paired \imode{} tasks through Knowledge Integration and Requirement Perturbation while preserving executable targets. We further design a controlled \ssprotocol{} protocol that allows models to clarify user intent and inspect the database environment without exposing hidden execution feedback. Evaluating seven language models, we find that the best average scores are only 62.2\% and 57.8\% in \dmode{} and \imode{}, respectively, demonstrating that realistic NL-to-PL/SQL development remains challenging, particularly in interactive settings. We release the benchmark on the public website \url{https://zeyuhuangzh.github.io/procarena-site/}.

\end{abstract}

\maketitle

\vldbtopmatter

\input{contents/introduction}
\input{contents/preliminaries}
\input{contents/benchmark_construction}
\input{contents/statistics}
\input{contents/experiments}
\input{contents/related_work}
\input{contents/conclusion}

\bibliographystyle{ACM-Reference-Format}
\bibliography{sample}

\end{document}

%% file: contents/introduction.tex
\section{Introduction}

\begin{figure*}[t]
    \centering
    \includegraphics[width=\textwidth]{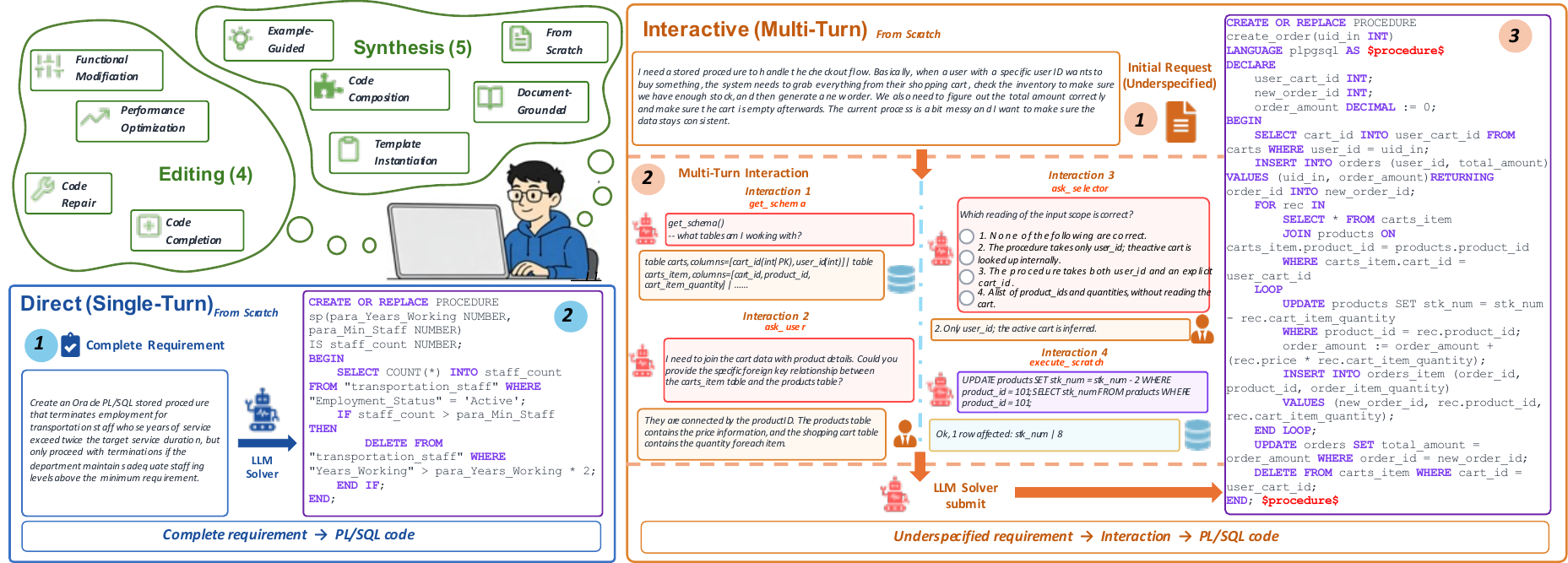}
    \Description{On the left, nine PL/SQL development subscenarios are grouped into five Synthesis
    subscenarios and four Editing subscenarios,
    below which a Direct episode passes a complete requirement to the Solver,
    which returns Oracle PL/SQL code in one turn.
    On the right, an Interactive episode starts from an underspecified checkout-flow request,
    proceeds through four numbered turns that propose an interpretation, ask for a foreign-key relationship,
    propose an execution sequence, and execute a scratch statement against the database,
    and ends with the Solver submitting a PL/pgSQL procedure.}
    \caption{\textbf{\benchmark{} evaluates the same PL/SQL development target under two modes.}
    Left: the nine subscenarios, five under Synthesis and four under Editing,
    and a \dmode{} episode that maps a complete requirement to PL/SQL code in a single turn.
    Right: an \imode{} episode that clarifies an underspecified request via \texttt{ask\_selector} and \texttt{ask\_user} turns with the \emph{User}
    , \texttt{execute} turns against the \emph{Database Environment} before the code is submitted.}
    \label{fig:overview}
\end{figure*}

By integrating procedural constructs with SQL, PL/SQL\footnote{In this paper, we use PL/SQL to refer to procedural extensions of SQL, including Oracle PL/SQL~\cite{oracleDoc}, PostgreSQL PL/pgSQL~\cite{postgresqlDoc}, and related dialects.} enables developers to implement data access and business logic directly within database systems, reducing communication overhead between applications and databases~\cite{hu2024webridge}.
It has therefore become an important component of modern database systems~\cite{de2025effectiveness,feuerstein2005oracle,douglas2003postgresql,ben2009inside}.
For example, Microsoft Azure SQL Database alone hosts over two billion PL/SQL codes that serve billions of daily invocations~\cite{microsoft_azure_sql,gupta2021procedural}.
However, developing PL/SQL codes is difficult in practice, as it requires knowledge of database schemas, business logic, and PL/SQL syntax~\cite{pavlo2017we, hu2024webridge}. Recently, large language models (LLMs) have been increasingly explored for automating the translation from natural language (NL) to PL/SQL (NL-to-PL/SQL)~\cite{zhang2025plforge}.

Despite this progress, systematically assessing LLMs for practical NL-to-PL/SQL development remains challenging, as no benchmark captures the diverse development scenarios and interactions involved in real-world PL/SQL programming. Prior work provides only partial foundations for such an assessment. PLForge~\cite{zhang2025plforge} introduces NL-to-PL/SQL datasets, models, and execution-based evaluation, but primarily studies direct PL/SQL code generation from complete requirements. NL-to-SQL benchmarks~\cite{yu2018spider,li2023bird,10.1145/3737873,lei2025spider} examine how models translate NL requests into SQL by grounding them in database schemas and contents, while interactive Text-to-SQL benchmarks such as BIRD-Interact~\cite{huo2026bird} further incorporate interactions with users and database environments. However, both lines of work target declarative SQL rather than PL/SQL codes. General code-generation benchmarks~\cite{chen2021evaluating,raihan2025mhumaneval,jain2025livecodebench,lu2021codexglue} cover generation and modification tasks in languages such as Python, Java, and C++, but do not capture key characteristics of PL/SQL development, including database environment and procedural logic.

To the best of our knowledge, \benchmark{} is the first benchmark to evaluate LLMs on NL-to-PL/SQL across multiple development scenarios in both \dmode{} and \imode{} modes across PostgreSQL and Oracle. Figure~\ref{fig:overview} illustrates the \dmode{} and \imode{} task under from scratch development scnario. To construct the benchmark, we outline two main challenges. \textbf{C1: How can we construct realistic and difficult NL-to-PL/SQL tasks across multiple PL/SQL development scenarios?}
Existing publicly available NL-to-PL/SQL datasets rarely cover these diverse development scenarios. They primarily focus on from-scratch generation with complete NL requirements, resulting in relatively limited scenario diversity and task difficulty. \textbf{C2: How can we construct an executable interactive environment for realistic multi-turn NL-to-PL/SQL development?}
Interactive NL-to-PL/SQL development requires the Solver to resolve incomplete requirements against both user intent and the database environment. A fixed transcript cannot assess whether the Solver appropriately seeks clarification or inspects database information, while unrestricted interaction may alter database state or leak evaluation signals. The environment must therefore support controlled interactions between the Solver and both the User and the Database Environment under an executable protocol.

For C1, we use iterative logic enhancement to add procedural logic, jointly update the PL/SQL code and NL requirement, and validate their executability and semantic consistency.
We then apply scenario-specific adapters to construct executable \dmode{} tasks for nine subscenarios.
For each subscenario, we rewrite the user requirement and prepare any required supporting artifact, such as a document, behavioral examples, a template, or existing code, while preserving the gold PL/SQL code.
For C2, we apply knowledge injection and requirement perturbation to derive paired \imode{} tasks from their \dmode{} counterparts by introducing information omissions and resolvable contradictions while preserving the gold PL/SQL code. We maintain an elicitation ledger of the facts, definitions, and resolutions available to the \usersimulator{} during interaction. We further design the \ssprotocol{} protocol to coordinate user interactions with database environment.

We propose \benchmark{}, which comprises 3{,}998 NL-to-PL/SQL tasks across nine PL/SQL development scenarios in PostgreSQL and Oracle.
We evaluate 7 models across \dmode{} and \imode{} tasks.
On the \dmode{} and \imode{} modes, the best average scores achieved by the models are 62.2\% and 57.8\%, respectively.
The evaluation further compares generation across subscenarios and analyzes the construction components and User Simulator.

In summary, we make the following contributions:
\begin{enumerate}
    \item We propose \benchmark{}, the first benchmark to evaluate LLMs on NL-to-PL/SQL across multiple development scenarios in both \dmode{} and \imode{} modes across PostgreSQL and Oracle.
    \item We construct PL/SQL-specific interactive environments through knowledge integration and requirement perturbation, and propose the \ssprotocol{} protocol to coordinate Solver--User interactions and Database Environment tools.
    \item We conduct comprehensive experiments to evaluate seven models on 3,998 tasks across nine development scenarios in both \dmode{} and \imode{} modes, and further analyze the benchmark construction and interaction behavior.
\end{enumerate}

%% file: contents/preliminaries.tex
\section{Problem Formulation}
\label{sec:formulation}

An \nlplsql{} task asks a \solver{} to produce PL/SQL code
from an NL requirement,
a database description,
and an optional task attachment.
We represent the task input as
\begin{equation}
\label{eq:task}
X = \langle q,\ \mathcal{D},\ \mathcal{A} \rangle,
\end{equation}
where $q$ states the NL requirement.
The database description
$\mathcal{D} = \langle \mathcal{D}_s,\, \mathcal{D}_m \rangle$
contains the database schema $\mathcal{D}_s$
and its metadata $\mathcal{D}_m$.
$\mathcal{A}$ denotes the optional task attachment.

Let $\mathbb{P}$ denote the space of PL/SQL codes.
Given a task input $X$, the \solver{} ultimately produces
a candidate PL/SQL code $\hat{P} \in \mathbb{P}$.
We model the \solver{} as
\begin{equation}
\mathcal{S}
= \langle \mathcal{M},\ \mathbb{A} \rangle ,
\end{equation}
where $\mathcal{M}$ is the LLM
and $\mathbb{A}$ is its action space.

\subsection{Direct NL-to-PL/SQL}
\label{subsec:direct-formulation}

The direct \nlplsql{} task provides a task
$X_d = \langle q_d,\mathcal{D}_d,\mathcal{A}_d \rangle$
to a \solver{} $\mathcal{S}_d
= \langle \mathcal{M}_d,\mathbb{A}_d \rangle$, where $\mathbb{A}_d = \{\texttt{submit}\}$.
The LLM directly generates a candidate PL/SQL code
$\hat{P}_d \in \mathbb{P}$ from the task input:
\begin{equation}
\hat{P}_d = \mathcal{M}_d(X_d) .
\end{equation}

Generation follows $q_d$ under $\mathcal{D}_d$
and may use the task attachment $\mathcal{A}_d$.
The \solver{} then submits $\hat{P}_d$ for evaluation.

\subsection{Interactive NL-to-PL/SQL}
\label{subsec:interactive-formulation}

An interactive NL-to-PL/SQL task is a multi-turn collaboration
among a \solver{} $\mathcal{S}_i$,
a \usersimulator{} $\mathcal{U}_i$,
and a database environment $\mathcal{E}_i$.
The $\mathcal{S}_i$ receives the initial task input from $\mathcal{U}_i$:
\begin{equation}
X_i = \langle q_i,\ \mathcal{D}_i,\ \mathcal{A}_i \rangle ,
\end{equation}
where $q_i$ may be underspecified.
For a paired ProcArena task,
$\mathcal{D}_i=\mathcal{D}_d$ and $\mathcal{A}_i=\mathcal{A}_d$;
interactive construction changes the requirement view while preserving the database description and task attachment.
The $\mathcal{S}_i$ can clarify information through the $\mathcal{U}_i$
and inspect the database through $\mathcal{E}_i$.

The interaction action space $\mathbb{A}_i$ of $\mathcal{S}_i$ is
\begin{equation}
\mathbb{A}_i
= \mathbb{A}_{\mathcal{U}}
\cup
\mathbb{A}_{\mathcal{E}}
\cup
\{\texttt{submit}\},
\end{equation}
where $\mathbb{A}_{\mathcal{U}}$ and $\mathbb{A}_{\mathcal{E}}$ denote the action spaces for interacting with $\mathcal{U}_i$ and $\mathcal{E}_i$, respectively.

The \solver{} $\mathcal{S}_i$ initiates each interaction turn.
Starting from an empty interaction history $h_0=\emptyset$,
at an interaction turn $t$, the LLM $\mathcal{M}_i$
selects an action based on the initial task input $X_i$
and the interaction history $h_{t-1}$:
\begin{equation}
\label{eq:episode-action}
a_t = \mathcal{M}_i(X_i,h_{t-1}),
\qquad
a_t \in \mathbb{A}_i .
\end{equation}

For a nonterminal action,
the corresponding interaction target returns a response
\begin{equation}
r_t =
\mathcal{R}
(a_t,h_{t-1};\mathcal{U}_i,\mathcal{E}_i),
\end{equation}
and the protocol updates the interaction history as
\begin{equation}
h_t =
h_{t-1} \oplus \langle a_t,r_t \rangle ,
\end{equation}
where $\oplus$ denotes appending the current action--response pair
to the interaction history.

After zero or more nonterminal interaction turns,
the LLM generates a candidate PL/SQL code
$\hat{P}_i \in \mathbb{P}$ based on the initial task input
and the interaction history:
\begin{equation}
\hat{P}_i = \mathcal{M}_i(X_i,h_T),
\end{equation}
where $T$ is the number of interaction turns.
The \solver{} $\mathcal{S}_i$ then submits $\hat{P}_i$ for evaluation,
which terminates the task.

%% file: contents/benchmark_construction.tex
\section{Benchmark Construction}
\label{sec:construction}

\begin{figure}[t]
    \centering
    \includegraphics[width=0.85\columnwidth]{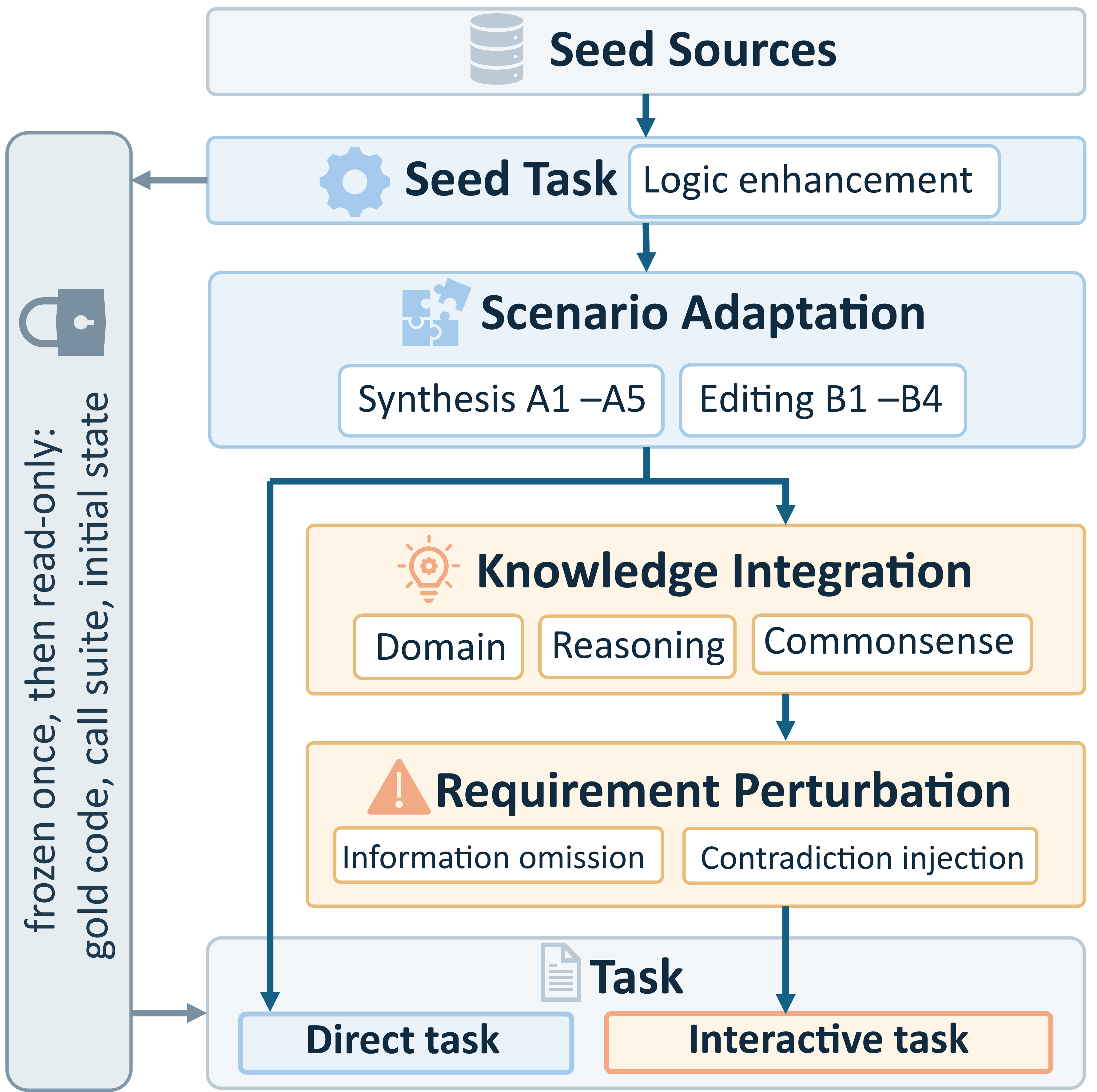}
    \caption{The \benchmark{} construction pipeline.}
    \label{fig:construction}
\end{figure}

In this section, we describe the construction of \benchmark{} tasks.
Figure~\ref{fig:construction} illustrates the overall construction pipeline.
We first construct the seed PL/SQL codes and enhance their procedural logic through an iterative logic enhancement method. We then rewrite the corresponding NL descriptions to preserve semantic consistency with the enhanced PL/SQL codes. For direct NL-to-PL/SQL task construction, we employ scenario adapters to transform the enhanced NL-to-PL/SQL pairs into direct NL-to-PL/SQL tasks spanning nine development scenarios. For interactive NL-to-PL/SQL task construction, we apply knowledge injection and requirement perturbation to transform the direct NL-to-PL/SQL tasks into interactive NL-to-PL/SQL tasks that necessitate interaction between the \solver{} and the user to achieve correct task completion.

\subsection{Seed Task Construction}
\label{subsec:reference-construction}

We construct the seed NL-to-PL/SQL tasks from two sources. The first consists of the Hard and Webridge datasets from PLForge, whose underlying databases are derived from Spider. These datasets provide difficult NL-to-PL/SQL pairs in PL/pgSQL and Oracle PL/SQL, respectively. The second source is LogicCat~\cite{mao2026logiccat}, which contains databases and NL-to-SQL tasks that require more sophisticated logical reasoning and domain knowledge. Using the LogicCat databases together with their original complex NL-to-SQL examples as prompts, we employ LLMs to generate corresponding seed NL-to-PL/SQL pairs.

\subsubsection{Iterative Logic Enhancement}
\label{subsubsec:logic}

Due to the strong code generation capabilities of state-of-the-art LLMs, we need to ensure that the NL-to-PL/SQL pairs in \benchmark{} are sufficiently challenging to effectively evaluate the capabilities of LLMs. In practice, users can extend a requirement by adding business rules. We leverage this characteristic and propose an iterative logic enhancement method, which let a held-out LLM to add one reasonable business rule in each round. Through iterative enhancement, we increase the procedural complexity of the PL/SQL code while ensuring that it remains grounded in realistic business scenarios.

Concretely, the iterative procedure starts from an NL-to-PL/SQL pair in the source set and proceeds over multiple rounds. In each round, given the current NL-to-PL/SQL pair, the held-out LLM proposes a business rule, generates a PL/SQL code snippet that implements the rule, integrates the snippet into the current PL/SQL code, and updates the NL requirement to describe the resulting PL/SQL code.After $N$ rounds, we execute the final PL/SQL code on the database and use a held-out validation LLM to check whether the NL requirement and code remain semantically aligned.
We then generate a call suite and retain only calls that change the database state.
The validated pair and its execution assets are collectively referred to as the seed task for subsequent \dmode{} and \imode{} task construction. We illustrate the iterative logic enhancement method in Algorithm~\ref{alg_logic_enhance}.

\begin{algorithm}[t]
\caption{Iterative Logic Enhancement}
\label{alg_logic_enhance}
\DontPrintSemicolon
\SetAlgoNoEnd
\SetKwInOut{Input}{Input}
\SetKwInOut{Output}{Output}

\Input{Source pair $\langle q_0,P_0\rangle$; database description $\mathcal{D}$; initial state $\sigma_0$; held-out construction LLM $\mathcal{M}_c$; enhancement rounds $N$; repair budget $B$}
\Output{Validated seed task $\tau^{+}$, or \textsc{Fail}}
$(q,P)\gets(q_0,P_0)$\;
\tcc{Add one rule per round}
\For{$n\gets1$ \KwTo $N$}{
  $(\hat q,\hat P)\gets\textsc{Enhance}(\mathcal{M}_c,q,P,\mathcal{D})$\;
  $(q,P)\gets(\hat q,\hat P)$\;
}
$\eta\gets\textsc{Execute}(P,\mathcal{D},\sigma_0)$\;
\If{$\textsc{Failed}(\eta)$}{
  \tcc{Repair from execution feedback}
  \For{$b\gets1$ \KwTo $B$}{
    $P\gets\textsc{Repair}(\mathcal{M}_c,q,P,\mathcal{D},\eta)$\;
    $\eta\gets\textsc{Execute}(P,\mathcal{D},\sigma_0)$\;
    \If{$\textsc{Succeeded}(\eta)$}{\textbf{break}\;}
  }
}
\If{$\textsc{Failed}(\eta)$ or $\textsc{CheckAlignment}(\mathcal{M}_c,q,P,\mathcal{D})=\textbf{false}$}{\Return{\textsc{Fail}}}
$\hat C\gets\textsc{GenerateCalls}(\mathcal{M}_c,q,P,\mathcal{D})$\;
$C\gets\textsc{KeepStateChanging}(\hat C,P,\mathcal{D},\sigma_0)$\;
\If{$C=\emptyset$}{\Return{\textsc{Fail}}}
\Return{$\tau^{+}=\langle q,\mathcal{D},\emptyset,\sigma_0,P,C\rangle$}\;
\end{algorithm}

\subsection{\dmode{} Task Construction}
\label{subsec:scenarios}
\label{subsec:direct}

In this section, we first define the scenarios and then explain how to construct \dmode{} tasks based on the seed tasks.

\subsubsection{Development Scenarios}
\label{subsubsec:development-scenarios}

We collect 1,535 unique posts from Stack Overflow and Database Administrators Stack Exchange by searching for keywords such as PL/SQL and PL/pgSQL. We then analyze the types of user requests in these posts and find that most could be broadly categorized into two categories. In 48.7\% of the posts, users have no existing PL/SQL code and need to generate code from scratch. In another 38.4\%, users have PL/SQL code but need to modify, optimize, fix, or complete it. We refer to these two categories as \textit{Synthesis} and \textit{Editing}, respectively. Based on the analysis, we further divide \textit{Synthesis} and \textit{Editing} into five and four subscenarios, respectively.

For \textit{Synthesis}, we define five subscenarios according to the information or code artifact provided with the task:
\begin{itemize}
    \item \emph{A1: From-Scratch Synthesis}.
    The user provides an NL requirement and the database description and requests complete target code from scratch.
    \item \emph{A2: Document-Grounded Synthesis}.
    The user also provides a design or requirement document that the code must follow.
    \item \emph{A3: Example-Guided Synthesis}.
    The user provides examples that illustrate the expected behavior of the target code.
    \item \emph{A4: Template Instantiation}.
    The user provides a generic code template and requests a target implementation that fills its task-specific bindings.
    \item \emph{A5: Code Composition}.
    The user provides complete, callable PL/SQL components and requests new target code that coordinates their calls,
    parameters,
    and procedural logic.
\end{itemize}

For \textit{Editing}, we define four subscenarios based on the type of requested code change:
\begin{itemize}
    \item \emph{B1: Functional Modification}.
    The user requests changes that add,
    remove,
    or modify business behavior in existing code.
    \item \emph{B2: Performance Optimization}.
    The user requests better execution performance while preserving the behavior of existing PL/SQL code.
    \item \emph{B3: Code Repair}.
    The user reports a defect in PL/SQL code and requests a repair that restores the intended behavior.
    \item \emph{B4: Code Completion}.
    The user provides incomplete PL/SQL code and requests completion of a coherent missing block.
\end{itemize}

\subsubsection{Scenario-Specific Adapters}
\label{subsubsec:scenario-adapters}

We use scenario-specific adapters driven by held-out LLM to transform a validated seed task into a \dmode{} task for each subscenario.
The validated seed task defines the target behavior.
Each scenario adapter creates the starting point for one development subscenario by rewriting the user requirement and adding a task attachment when needed.
The attachment may contain a document,
examples,
a template,
reusable components,
or existing PL/SQL code.
Let $\tau^{+}$ denote the validated seed task.
The scenario contract $G_j$ defines the preconditions under which adapter $F_j$ applies and the acceptance criteria that its output must satisfy.
For subscenario $j$, the adapter constructs the complete \dmode{} task
\begin{equation}
\label{eq:scenario-adapter}
F_j(\tau^{+};G_j)
=
\tau_{d,j}
=
\langle q_{d,j},\mathcal{D},\mathcal{A}_j,\sigma_0,P,C\rangle.
\end{equation}
Here, $q_{d,j}$ denotes the scenario-specific requirement, and $\mathcal{A}_j$ denotes the task attachment.
The remaining elements correspond to the database description, initial database state, reference PL/SQL code, and call suite, all of which remain unchanged.

Each scenario adapter constructs a task by rewriting the original requirement into a scenario-specific requirement and, when needed, generating a task attachment, while leaving all other task elements unchanged.
The resulting NL requirement and attachment must contain all information necessary to complete the task and remain consistent with the database description.
If candidate content contains conflicting statements, the adapter resolves the conflict and retains a single consistent statement.

To construct the scenario-specific requirement and attachment, the adapter combines LLM-guided rewriting with scenario-specific artifact preparation.
The LLM handles content that requires semantic rewriting.
For Synthesis subscenarios that require an attachment, the adapter places the appropriate supporting artifact in the attachment: a document, behavioral examples, the provided generic template, or callable components.
For Editing subscenarios, deterministic transformations derive code attachments from the reference PL/SQL code without modifying the reference code itself.
These transformations revert a business rule,
introduce a behavior-preserving slowdown,
inject a defect,
or remove a coherent code block.
After each attempt, we validate the candidate against its scenario contract.
If validation fails, the LLM receives the failed checks as feedback for the next attempt.

Every candidate must pass three common checks to ensure that the constructed \dmode{} task is valid, non-trivial, and semantically aligned with the reference seed task.
\ding{172} \emph{Artifact validity} verifies that the attachment has the type and structure required by the target subscenario.
\ding{173} \emph{Task non-triviality} ensures that the candidate neither exposes the reference code nor leaves no meaningful work for the \solver{}.
\ding{174} \emph{Reference alignment} verifies semantic consistency with the reference code through execution.

In addition to these common checks, each candidate must pass the scenario-specific checks defined in Table~\ref{tab:scenario-adapters}.

\begin{table*}[t]
\captionsetup{font=large}
\caption{Scenario-specific adapters.
A1--A5 cover Synthesis and B1--B4 cover Editing.
All adapters preserve the database description,
reference code,
call suite,
and initial database state.
Each constructed user input is written as requirement; \artifact{attachment}.
The last column lists the checks applied after the three common checks.}
\label{tab:scenario-adapters}
\centering
\scriptsize
\setlength{\tabcolsep}{4pt}
\setlength{\extrarowheight}{6pt}
\begin{tabularx}{\textwidth}{@{}>{\raggedright\arraybackslash}p{0.16\textwidth}R{0.86}R{0.90}R{1.24}@{}}
\toprule
\textbf{Subscenario} & \textbf{Applicable seed task} & \textbf{Constructed user input} & \textbf{Additional acceptance criteria} \\
\midrule
\textbf{A1: From-Scratch}
& Any validated seed task.
& Complete behavioral requirement; \artifact{no attachment}.
& \ding{182}~The task has no attachment. \ding{183}~The requirement fully specifies the target behavior. \\
\zebra \textbf{A2: Document-Grounded}
& The target behavior depends on facts that can form a separate document.
& Task request without the document facts; \artifact{document stating them}.
& \ding{182}~Some implementation that ignores a document fact differs from the reference under the call suite. \ding{183}~The document exposes no code. \\
\textbf{A3: Example-Guided}
& The call suite contains inputs that illustrate the target behavior.
& Task request; \artifact{calls selected from the call suite with their observed behavior}.
& \ding{182}~The reference code reproduces every example. \ding{183}~A leakage check rejects PL/SQL implementation text. \\
\zebra \textbf{A4: Template Instantiation}
& The target admits a reusable skeleton and task-specific bindings.
& Task-specific bindings; \artifact{reusable skeleton with unfilled placeholders}.
& \ding{182}~A placeholder-coverage check matches every binding to the template. \ding{183}~The check rejects a target-specific body. \\
\textbf{A5: Code Composition}
& Existing callable components can implement the target when coordinated.
& Composition request; \artifact{callable PL/SQL components without the coordination code}.
& \ding{182}~The components compile and are callable. \ding{183}~A structural check confirms the target coordination code is absent. \\
\midrule
\textbf{B1: Functional Modification}
& Reference code has a business rule that functional reversion can invert.
& Original rule as the requested change; \artifact{reference code with that rule reverted}.
& \ding{182}~At least one call distinguishes the unchanged code from the reference behavior. \\
\zebra \textbf{B2: Performance Optimization}
& A behavior-preserving slowdown operator applies to the target.
& Optimization request; \artifact{behavior-preserving slowdown of the reference code}.
& \ding{182}~The slower code matches the reference outputs and state. \ding{183}~Its warm median runtime exceeds the reference by the contract margin. \\
\textbf{B3: Code Repair}
& A defect operator applies to the working code.
& Repair request; \artifact{reference code with an injected defect}.
& \ding{182}~At least one call exposes the defect: an exception, or a mismatched output or persistent state. \\
\zebra \textbf{B4: Code Completion}
& The reference code contains a coherent removable block.
& Completion request; \artifact{reference code with a coherent block removed and the site marked}.
& \ding{182}~A structural check validates the removed block. \ding{183}~Compilation or execution confirms the incomplete code fails the target behavior. \\
\bottomrule
\end{tabularx}
\end{table*}

After an adapter passes all checks, we materialize the resulting \dmode{} task.
The \solver{} receives
\begin{equation}
X_{d,j}=\langle q_{d,j},\mathcal{D},\mathcal{A}_j\rangle.
\end{equation}
This input contains only the scenario-specific requirement,
database description,
and task attachment.
We keep the gold PL/SQL code,
call suite,
and initial database state hidden for offline evaluation.
In \dmode{}, the \solver{} is allowed to submit only once and cannot call \texttt{ask\_user}, \texttt{ask\_selector}, or any environment tool before submission.
We construct \dmode{} tasks for all nine subscenarios.

\subsection{\imode{} Task Construction}
\label{subsec:interactive}

\subsubsection{Interactive Requirement Construction}
\label{subsubsec:interactive-requirements}
\label{subsubsec:knowledge}

\begin{figure}[t]
    \centering
    \includegraphics[width=\columnwidth]{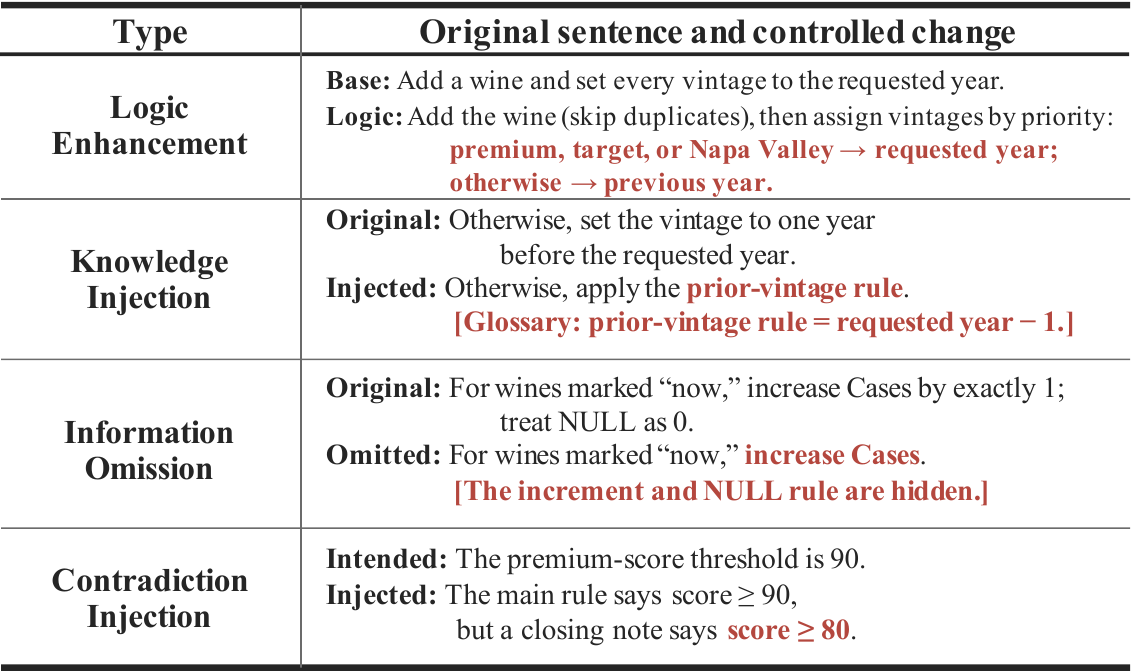}
    \caption{Controlled requirement edits on one example task.
    Each row pairs the original sentence with the rewritten one produced by a single operator,
    with the changed span highlighted:
    Iterative Logic Enhancement, Knowledge Injection, Information Omission, and Contradiction Injection.}
    \label{fig:example}
\end{figure}

We construct \imode{} tasks from the \dmode{} tasks in Section~\ref{subsec:direct} through two stages.
In real-world development, user requirements are often not fully explicit: users may express their intent using domain-specific knowledge, logical reasoning, or commonsense knowledge, and may also omit necessary information or provide conflicting constraints.
We model these realistic forms of underspecification through Knowledge Injection and Requirement Perturbation.
Figure~\ref{fig:example} illustrates both stages on one airline-operations task,
together with the Iterative Logic Enhancement step that produces the underlying \dmode{} requirement.

Both stages modify only the requirement and preserve the task attachment, reference PL/SQL code, call suite, and initial database state.
We apply this construction to eight subscenarios; B2 remains \dmode{}-only because it uses runtime-based evaluation.

\textbf{Knowledge Injection} transforms a complete \dmode{} requirement by replacing selected explicit statements with formulations that rely on implicit knowledge.
It uses three operators:
\begin{itemize}
    \item \emph{Domain Knowledge} replaces an expanded definition or formula with a domain term.
    \item \emph{Logical Reasoning} replaces procedural steps with an invariant or constraint.
    \item \emph{Commonsense Knowledge} removes information that users infer from common conventions in the given scenario.
\end{itemize}
All three operators preserve the scenario contract and the reference PL/SQL code.

For all three operators, we extract the knowledge needed to interpret each rewrite and store it in the online knowledge ledger $\Lambda_j^K$.
Each entry contains a domain definition, reasoning logic, or commonsense knowledge and is available to the \solver{} during interaction.

Algorithm~\ref{alg_knowledge_injection} summarizes the Knowledge Injection process.
For each rewritten requirement fragment produced by Knowledge Injection, the algorithm records its knowledge type, the corresponding source fact, the rewritten text, and any queryable knowledge entry.
It then validates the rewritten fragment by checking whether it preserves the scenario characteristics, remains consistent with the gold PL/SQL code, avoids schema-identifier leakage, provides sufficient ledger coverage, and can be traced back to the source fact.
If a rewritten fragment fails any validation check, a new rewrite is generated within the fixed attempt budget.

\begin{algorithm}[t]
\caption{Knowledge Injection}
\label{alg_knowledge_injection}
\DontPrintSemicolon
\SetAlgoNoEnd
\SetKwInOut{Input}{Input}
\SetKwInOut{Output}{Output}

\Input{Task $\tau_{d,j}$; contract $G_j$; knowledge budget $\kappa$; attempt budget $A$}
\Output{Interactive requirement candidate $q_{i,j}^{K}$ and ledger $\Lambda_j^{K}$, or \textsc{Fail}}
$\Pi\gets\textsc{EditableProse}(q_{d,j},G_j)$; $\mathit{fb}\gets\emptyset$\;
\If{$\Pi=\emptyset$}{\Return{\textsc{Fail}}}
\For{$a\gets1$ \KwTo $A$}{
  $(\hat\Pi,\hat\Lambda^K)\gets\textsc{LLMInject}(\Pi,P,\mathcal{D},G_j,\kappa,\mathit{fb})$\;
  $\mathit{fb}\gets\textsc{KnowledgeGates}(\hat\Pi,\hat\Lambda^K;\Pi,P,\mathcal{D},G_j)$\;
  \If{$\mathit{fb}=\emptyset$}{
    $q_{i,j}^{K}\gets\textsc{ApplyRewrite}(q_{d,j},\Pi,\hat\Pi)$\;
    \Return{$q_{i,j}^{K},\hat\Lambda^K$}\;
  }
}
\Return{\textsc{Fail}}\;
\end{algorithm}

\label{subsubsec:perturbation}

\textbf{Requirement Perturbation} applies two operators to $q_{i,j}^{K}$ after Knowledge Injection.
\begin{itemize}
    \item \emph{Information Omission} removes necessary but recoverable information from the visible requirement.
    \item \emph{Contradiction Injection} introduces conflicts within the requirement or between the requirement and the database environment.
\end{itemize}

\label{subsubsec:omission}
\emph{Information Omission} removes an explicit but recoverable requirement detail from the visible requirement.
We regard the removed detail as necessary if its omission permits an implementation that satisfies the remaining requirement but behaves differently from the reference PL/SQL code when both are executed on the same call suite and initial database state.
For each omitted detail, the requirement ledger records the original statement, where it appears in the source requirement, who can provide the missing information, and how the detail can be restored during interaction.

\label{subsubsec:contradiction}
\emph{Contradiction Injection} introduces conflicting information that cannot be resolved from the visible requirement alone. We consider two forms of contradiction.

An \emph{Intra-requirement contradiction} introduces two incompatible statements into the same requirement, such as different thresholds. Because both statements are visible, the \solver{} can detect the conflict directly from the requirement, but must interact with the user to determine which statement the user actually intends.

A \emph{Requirement-environment contradiction} introduces a statement that conflicts with the actual database schema or metadata. For example, the requirement may refer to a nonexistent table or column, or specify an incorrect data type or constraint. The \solver{} can use environment tools to detect such mismatches. Environment tools can detect the mismatch, but cannot determine the user's intended correction. The conflict therefore requires interaction with the user for resolution.

Both perturbation operators modify only the visible requirement while preserving the database schema, metadata, gold PL/SQL code, call suite, and initial database state.
Each injected conflict has a unique resolution that is consistent with the gold PL/SQL code. For requirement-environment contradictions, we additionally require that the mismatch can be detected from the database environment.

\subsubsection{Task Materialization and Interaction Protocol}
\label{subsubsec:interactive-environment}

\textbf{Task Materialization.}
After Knowledge Injection and Requirement Perturbation, we package the resulting requirement and its supporting assets into an \imode{} task:
\begin{equation}
\tau_{i,j}
=
\langle q_{i,j},\mathcal{D},\mathcal{A}_j,\sigma_0,P,C,\Lambda_{i,j}\rangle.
\end{equation}
The \solver{} receives only the requirement, database description, and task attachment.
The router assigns each ledger entry to its designated information holder.
The pipeline retains only tasks whose ledger entries identify the corresponding construction operators.

\label{subsec:protocol}

\textbf{Interaction Protocol.}
The \usersimulator{} represents the user in each interactive episode and responds to the \solver{}'s questions and clarification requests. Since the interaction proceeds over multiple turns in free-form natural language, we implement the simulator with a held-out LLM.
The \ssprotocol{} protocol governs how the \solver{} interacts with the simulator. It defines the actions available to the \solver{}, the task information that the simulator may access for each action, and the responses that the simulator may provide. By restricting the simulator to user-facing information, the protocol prevents hidden evaluation assets from leaking into the interaction.
Within the simulator, the requirement holder and knowledge holder handle requests that match entries in their corresponding ledgers.

During task construction, we create an NL description of the gold PL/SQL behavior for the fallback channel, which answers questions without a matching ledger entry. Separately, environment tools expose the database environment defined in Section~\ref{subsec:interactive-formulation}, allowing the \solver{} to inspect task-relevant database information through the prescribed tool interface.

\textbf{Action Space.}
The router sends each action to the component selected by the \solver{}, and the offline evaluator remains outside the interaction.
The \solver{} is the only action initiator in the protocol.
It may inspect the environment, request information through \texttt{ask\_user}, ask the user to select among candidate interpretations through \texttt{ask\_selector}, or end the episode through \texttt{submit}.
Table~\ref{tab:actions} summarizes the nine concrete actions, their target components, inputs, returned observations, and interaction costs.
The two user-facing actions return a natural-language answer and the index of the intended option, respectively.
The analysis groups malformed or unrecognized names under \texttt{malformed/other}; this category falls outside the action space.

\begin{table}[t]
\captionsetup{font=large}
\caption{Nine actions available to the \solver{} in an \imode{} episode.
Every issued action consumes budget units, even if malformed or failed.}
\label{tab:actions}
\centering
\scriptsize
\setlength{\tabcolsep}{3pt}
\begin{tabular}{@{}p{0.235\columnwidth}l p{0.165\columnwidth} p{0.335\columnwidth} r@{}}
\toprule
Action & Env. & Arguments & Returned observation & Cost \\
\midrule
\texttt{ask\_user} & User & \texttt{question} & Natural-language answer & 2 \\
\texttt{ask\_selector} & User & \texttt{options} & Index of the intended option & 2 \\
\midrule
\texttt{get\_schema} & DB & -- & Schema of the whole database & 1 \\
\texttt{describe\_table} & DB & \texttt{table} & Metadata of one table & 0.5 \\
\texttt{sample\_rows} & DB & \texttt{table}, \texttt{limit} & Sample rows of one table & 0.5 \\
\texttt{compile\_plsql} & DB & \texttt{code} & Compiler diagnostics & 0.5 \\
\texttt{execute\_scratch} & DB & \texttt{sql} & Execution result & 1 \\
\texttt{reset\_scratch} & DB & -- & Scratch context restored & 0.5 \\
\midrule
\texttt{submit} & Eval & \texttt{code} & - & 0 \\
\bottomrule
\end{tabular}
\end{table}

The task attachment $\mathcal{A}$ remains visible throughout the episode.
Every environment action and every \texttt{ask\_user} or \texttt{ask\_selector} action consumes the interaction budget;
\texttt{submit} is cost-free but terminates the episode immediately and returns no execution-accuracy feedback,
preventing the \solver{} from iteratively refining its solution against the hidden oracle.

The protocol provides two user-facing actions.
\texttt{ask\_user} routes the question to the holder responsible for the matched ledger entry, falling back to a default channel when no entry matches.
\texttt{ask\_selector} lets the \solver{} choose a holder and supply candidate interpretations; the holder returns the index consistent with its ledger.
When candidates mix requirement facts with injected knowledge, the \solver{} may query the two holders separately to resolve each source independently.

Resolving contradictions between the requirement and the database requires two complementary information sources.
The \solver{} first calls a schema or metadata action to identify the objective mismatch, such as a referenced column that does not exist,
then uses \texttt{ask\_user} or \texttt{ask\_selector} to obtain the user's intended resolution.
Environment actions expose database facts only and never infer user intent, ensuring a clean separation between objective state and subjective preference.

\FloatBarrier

%% file: contents/statistics.tex
\section{ProcArena Statistics}
\label{sec:stats}

\benchmark{} spans nine \dmode{} subscenarios and eight paired \imode{}
subscenarios across PostgreSQL and Oracle.
This breadth follows from the task itself: NL-to-PL/SQL development spans
synthesis and editing, complete and underspecified requirements, and two
dialects.
Varying the scenario, the mode, and the dialect therefore keeps evaluation
grounded in the full task rather than in one narrow setting.
This section reports the resulting corpus:
Table~\ref{tab:compare} summarizes its scale, coverage, and difficulty;
Table~\ref{tab:composition} breaks it down by subscenario; and
Table~\ref{tab:knowledge} quantifies the transformations that construct
\imode{} requirements.

\subsection{Comparison with Prior Datasets}
\label{sec:stats:compare}

Table~\ref{tab:compare} compares ProcArena with four prior NL-to-PL/SQL datasets:
PLForge Simple, PLForge Hard, ProcBench, and WeBridge~\cite{zhang2025plforge,hu2024webridge}.
\benchmark{} expands both scale and coverage.
The prior test splits contain 209--300 \dmode{} tasks, one dialect, one
scenario, and no \imode{} episodes.
\benchmark{} contains 3{,}998 tasks across two dialects and nine subscenarios,
including 1{,}879 \imode{} episodes.

\benchmark{} procedures also contain more statements and have higher
cyclomatic complexity on average.
They average 32.1 statements and 18.6 cyclomatic complexity, compared with
10.7 statements and 6.8 cyclomatic complexity for PLForge Hard and 29.5 and
9.4 for WeBridge.
\benchmark{} therefore covers more settings and has higher average procedure
complexity.

\begin{table}[t]
\centering
\captionsetup{font=large}
\caption{\benchmark{} and prior NL-to-PL/SQL datasets.
Tasks reports all evaluation instances, and Interactive reports the \imode{}
subset. Stmt and Cyclo report the mean statement count and cyclomatic
complexity per reference procedure.}
\label{tab:compare}
\scriptsize
\setlength{\tabcolsep}{3pt}
\begin{tabular*}{\columnwidth}{@{\extracolsep{\fill}}l ccccc cc@{}}
\toprule
& \multicolumn{5}{c}{Scale \& coverage} & \multicolumn{2}{c}{Per-task difficulty} \\
\cmidrule(lr){2-6}\cmidrule(lr){7-8}
Dataset & Tasks & DBs & Dialects & Scen. & Interactive & Stmt & Cyclo \\
\midrule
PLForge Simple & 300 & 130 & 1 & 1 & 0 & 2.8 & 3.0 \\
PLForge Hard   & 300 & 113 & 1 & 1 & 0 & 10.7 & 6.8 \\
ProcBench       & 300 & 123 & 1 & 1 & 0 & 3.3 & 3.5 \\
WeBridge        & 209 & 40  & 1 & 1 & 0 & 29.5 & 9.4 \\
\midrule
\textbf{ProcArena} & \textbf{3{,}998} & \textbf{157} & \textbf{2} & \textbf{9} & \textbf{1{,}879} & \textbf{32.1} & \textbf{18.6} \\
\bottomrule
\end{tabular*}
\end{table}

\subsection{Corpus Composition}
\label{sec:stats:composition}

Table~\ref{tab:composition} breaks down the corpus by subscenario.
\dmode{} task counts range from 111 to 120 per dialect.
\benchmark{} contains 1{,}060 PostgreSQL and 1{,}059 Oracle \dmode{} tasks.
Each \dmode{} task has a paired \imode{} episode except B2, which has no
\imode{} task because it uses a runtime score after semantic-equivalence
checking.
This pairing yields 940 PostgreSQL and 939 Oracle \imode{} episodes.
The corpus spans 157 databases; B3 uses 110, while A1 and B2 each use 76.

The Input rows report the mean size of the requirement and task attachment.
NL tokens include the requirement and any textual attachment, such as the A2
design document and A3 behavioral examples.
Code tokens count PL/SQL code in the attachment.
A1--A3 have no code attachment, so their Code-token cells are shown as dashes.

The Reference procedure rows summarize each target procedure and its call
suite: statements, lines, parameters, calls, and tables written.
Across the corpus, reference procedures average 39.0 lines, 4.0 parameters,
6.3 calls, and 4.9 tables written.

\begin{table}[t]
\centering
\captionsetup{font=large}
\caption{\benchmark{} corpus by subscenario. 
All gives the per-subscenario mean for task counts, the corpus-wide distinct
count for databases, and the per-applicable-task mean for other metrics.
A dash denotes a missing \imode{} task or a subscenario without a code attachment.}
\label{tab:composition}
\scriptsize
\setlength{\tabcolsep}{2.6pt}
\begin{tabular*}{\columnwidth}{@{\extracolsep{\fill}}l ccccc cccc c@{}}
\toprule
& \multicolumn{5}{c}{A. Synthesis} & \multicolumn{4}{c}{B. Editing} & \\
\cmidrule(lr){2-6}\cmidrule(lr){7-10}
Metric & A1 & A2 & A3 & A4 & A5 & B1 & B2 & B3 & B4 & All \\
\midrule
\multicolumn{11}{l}{\textit{Scale \& coverage}}\\
Direct, PG            & 114 & 119 & 118 & 119 & 111 & 119 & 120 & 120 & 120 & 117.8 \\
Direct, Oracle        & 115 & 116 & 120 & 116 & 118 & 118 & 120 & 120 & 116 & 117.7 \\
Interactive, PG       & 114 & 119 & 118 & 119 & 111 & 119 & -- & 120 & 120 & 117.5 \\
Interactive, Oracle   & 115 & 116 & 120 & 116 & 118 & 118 & -- & 120 & 116 & 117.4 \\
Databases             & 76  & 85  & 85  & 87  & 88  & 90  & 76  & 110 & 107 & 157 \\
\midrule
\multicolumn{11}{l}{\textit{Input (per task, avg)}}\\
NL tokens             & 609.5 & 898.5 & 764.3 & 1125.6 & 496.4 & 385.5 & 149.3 & 732.2 & 330.9 & 609.6 \\
Code tokens           & -- & -- & -- & 519.4 & 334.3 & 446.5 & 429.1 & 410.2 & 244.7 & 402.3 \\
\midrule
\multicolumn{11}{l}{\textit{Reference procedure (per task, avg)}}\\
Statements            & 36.9 & 31.2 & 43.6 & 40.7 & 22.7 & 41.9 & 15.0 & 26.7 & 30.5 & 32.1 \\
Lines                 & 39.8 & 45.4 & 54.4 & 39.2 & 45.1 & 54.5 & 13.3 & 32.7 & 26.7 & 39.0 \\
Parameters            & 4.0 & 4.0 & 4.1 & 3.6 & 4.9 & 3.9 & 3.9 & 3.9 & 4.1 & 4.0 \\
Calls $k$            & 6.4 & 6.1 & 6.4 & 6.4 & 6.7 & 6.4 & 5.2 & 6.3 & 6.5 & 6.3 \\
Tables written        & 5.1 & 4.8 & 5.8 & 5.9 & 4.2 & 5.8 & 2.3 & 4.6 & 5.2 & 4.9 \\
\bottomrule
\end{tabular*}
\end{table}

\subsection{Interactive Requirement Transformations}
\label{sec:stats:knowledge}

Table~\ref{tab:knowledge} reports six transformation types for \imode{}
requirement construction.
Section~\ref{subsubsec:interactive-requirements} groups them into Knowledge,
Contradiction, and Omission.
\emph{Knowledge} replaces explicit facts with domain terminology, reasoning
constraints, or commonsense conventions.
Domain definitions enter the elicitation ledger, while reasoning and
commonsense rewrites remain in the offline trace.
\emph{Contradiction} introduces incompatible readings within the requirement
or between the requirement and the environment.
\emph{Information Omission} removes a registered fact from the visible
requirement and stores it in the elicitation ledger.

\benchmark{} applies 5.75 transformations per \imode{} task on average.
A4 has fewer reasoning and commonsense rewrites than the other non-B3
subscenarios, while B3 has zero Knowledge transformations.
The construction pipeline changes only the requirement, so a subscenario can
retain transformations even when its attachment contains much of the target
code.

\begin{table}[t]
\centering
\captionsetup{font=large}
\caption{Interactive requirement transformations per task and subscenario.
B2 has no \imode{} task. Avg reports the mean across \imode{} tasks.}
\label{tab:knowledge}
\scriptsize
\setlength{\tabcolsep}{2.6pt}
\begin{tabular*}{\columnwidth}{@{\extracolsep{\fill}}l ccccc ccc c@{}}
\toprule
& \multicolumn{5}{c}{A. Synthesis} & \multicolumn{3}{c}{B. Editing} & \\
\cmidrule(lr){2-6}\cmidrule(lr){7-9}
Fact kind & A1 & A2 & A3 & A4 & A5 & B1 & B3 & B4 & Avg \\
\midrule
\multicolumn{10}{l}{\textit{Knowledge}}\\
Jargon (terminology)        & 1.89 & 3.01 & 2.05 & 1.00 & 1.56 & 1.81 & 0.00 & 1.07 & 1.54 \\
Reasoning                   & 0.84 & 0.78 & 0.86 & 0.43 & 0.76 & 0.84 & 0.00 & 0.85 & 0.67 \\
Commonsense                & 0.84 & 0.79 & 0.82 & 0.37 & 0.74 & 0.79 & 0.00 & 0.83 & 0.65 \\
\midrule
\multicolumn{10}{l}{\textit{Contradiction}}\\
Requirement                 & 0.68 & 0.69 & 0.68 & 0.67 & 0.68 & 0.69 & 0.68 & 0.68 & 0.68 \\
Environment                 & 0.32 & 0.31 & 0.32 & 0.33 & 0.32 & 0.31 & 0.33 & 0.32 & 0.32 \\
\midrule
\multicolumn{10}{l}{\textit{Omission}}\\
Omitted facts               & 2.00 & 2.60 & 2.00 & 1.99 & 1.58 & 1.94 & 2.00 & 1.00 & 1.89 \\
\midrule
Transformed facts           & 6.56 & 8.19 & 6.73 & 4.78 & 5.64 & 6.38 & 3.00 & 4.75 & 5.75 \\
\bottomrule
\end{tabular*}
\end{table}

%% file: contents/experiments.tex
\section{Experiments}
\label{sec:experiments}

We evaluate seven language models on \benchmark{},
across nine \dmode{} subscenarios, eight paired \imode{} subscenarios, and two SQL dialects.
The strongest model reaches 62.2\% execution accuracy in the \dmode{} mode
and 57.8\% in the \imode{} mode.
Sections~\ref{subsec:overall} through~\ref{subsec:error} report what the models achieve
and how they fail.
Section~\ref{subsec:validation} then checks that the task construction and the interactive
environment earn their place.

\subsection{Experimental Setup}
\label{subsec:setup}

\paragraph{Evaluation Modes.}
\benchmark{} presents all nine subscenarios in the \dmode{} mode
and eight of them in the \imode{} mode.
B2 Performance Optimization is \dmode{}-only,
because its score is execution time rather than state equality.
In the \dmode{} mode the \solver{} receives the complete requirement
and submits one procedure.
In the \imode{} mode it receives an underspecified requirement.
It may ask the \usersimulator{} for a fact with \texttt{ask\_user},
confirm a candidate reading with \texttt{ask\_selector},
and inspect the database with six environment tools before it submits.
Each paired task shares its database description,
task attachment,
gold procedure,
call suite,
and initial state.

\paragraph{Evaluation Metrics.}
Execution accuracy (EX) is the metric for the eight paired subscenarios in both modes.
We adapt PLForge's execution-match principle~\cite{zhang2025plforge}
to the stateful behavior of procedural code.
$\textsc{Exec}(P,C,\sigma_0)$ installs $P$ from the initial state $\sigma_0$,
runs the ordered call suite $C$,
and returns its outputs together with the final persistent state.
For a candidate $\hat P$,
the task-level execution indicator is
\begin{equation}
\label{eq:ex}
\operatorname{EX}(\hat{P};P,C,\sigma_0) =
\mathbf{1}\left[
\textsc{Exec}(\hat{P}, C, \sigma_0)
=
\textsc{Exec}(P, C, \sigma_0)
\right].
\end{equation}
Execution accuracy averages this indicator over tasks.

B2 is scored differently,
because its input is a degraded version of the gold procedure
and already satisfies Equation~\ref{eq:ex}.
Semantic equivalence is therefore a gate rather than a score.
An admitted candidate is measured against the gold procedure by two ratios,
\emph{acceleration} $T_{P}/T_{\hat P}$ and \emph{work} $W_{P}/W_{\hat P}$,
where $T$ is a median wall-clock time and $W$ a deterministic work counter.
Both are oriented so that $1.0$ means the candidate matched the gold procedure
and larger is better.
Section~\ref{subsec:perf} gives the measurement protocol.

For the \imode{} mode we additionally report the number of interaction turns
and the total tokens per episode.
Retention is defined in Section~\ref{subsec:interaction},
where it is first used.

\paragraph{Baseline Models.}
We evaluate seven language models:
GPT-5.5, GPT-5.4 Mini, and GPT-5.2~\cite{openai2025gpt5},
Gemini-3.1 Pro~\cite{google2025gemini3model},
GLM-5.3 and GLM-5.2~\cite{glm2026glm5},
and DeepSeek V4 Pro~\cite{deepseek2026v4}.

\paragraph{Implementation Details.}
Every task runs on a freshly restored database,
so no episode observes the writes of another.
\benchmark{} ships tasks in two dialects,
PL/pgSQL on PostgreSQL 12.22 and Oracle PL/SQL on Oracle Database 21c Express Edition.
A single fixed model, DeepSeek V4 Flash, drives the \usersimulator{} across every \solver{},
so an EX difference between two \solver{}s never reflects a difference in the simulator.
We set reasoning effort to high for every model that exposes it,
and we run each task once.
The \solver{} may take at most 60 recorded steps.
Its tool budget is $6 + 2|\Lambda| + 3$ units,
where $\Lambda$ is the task's elicitation ledger.
\texttt{ask\_user} and \texttt{ask\_selector} cost two units each,
environment tools cost between $0.5$ and one unit,
and \texttt{submit} is free.
Both the budget and the price list are shown to the \solver{},
and every tool result reports the units spent so far.

\begin{table*}[t]
\centering
\footnotesize
\setlength{\tabcolsep}{3pt}
\caption{\textbf{Interaction lowers accuracy on Synthesis and raises it on Editing.}
Execution accuracy (\%) per subscenario and dialect,
with \dmode{} above the shaded band and \imode{} below it.
Subscenarios run along each category's ordering axis of Section~\ref{subsec:scenarios}.
Within each mode block, the best value per column is in \textbf{bold}
and the second best is \underline{underlined}.}
\label{tab:main}
%
\small
\renewcommand{\arraystretch}{0.9}
\begin{tabular*}{\textwidth}{@{\extracolsep{\fill}}l ccccc ccc c ccccc ccc c@{}}
\toprule
& \multicolumn{9}{c}{\textbf{PostgreSQL}} & \multicolumn{9}{c}{\textbf{Oracle}} \\
\cmidrule(lr){2-10}\cmidrule(lr){11-19}
& \multicolumn{5}{c}{\textbf{A. Synthesis}} & \multicolumn{3}{c}{\textbf{B. Editing}} & \textbf{Overall}
& \multicolumn{5}{c}{\textbf{A. Synthesis}} & \multicolumn{3}{c}{\textbf{B. Editing}} & \textbf{Overall} \\
\cmidrule(lr){2-6}\cmidrule(lr){7-9}\cmidrule(lr){10-10}\cmidrule(lr){11-15}\cmidrule(lr){16-18}\cmidrule(lr){19-19}
Model & A1 & A2 & A3 & A4 & A5 & B1 & B3 & B4 & EX & A1 & A2 & A3 & A4 & A5 & B1 & B3 & B4 & EX \\
\midrule
\modeband{19}{\dmode{}}
\input{tables/direct_body.tex} 
\addlinespace[2pt]
\modeband{19}{\imode{}}
\input{tables/interactive_body.tex} 
\bottomrule
\end{tabular*}
\end{table*}

\subsection{Overall Results}
\label{subsec:overall}

Table~\ref{tab:main} reports EX for every model on every subscenario,
in both dialects.
The \dmode{} block sits above the shaded band and the \imode{} block below it.
The two blocks use the same paired tasks and the same row order,
so a reader compares them cell by cell.
B2 is not in this table;
Section~\ref{subsec:perf} reports it.

Gemini-3.1 Pro is the strongest model in both modes,
at 62.2\% and 57.8\% pooled over the two dialects.
GPT-5.5 follows at 61.6\% and 57.0\%.
DeepSeek V4 Pro is last in both, at 36.8\% and 23.7\%.
Even in the \dmode{} mode,
where the requirement is complete and no interaction is needed,
the best model is wrong on more than a third of the tasks.
The two dialects behave alike,
and their differences are small and not consistently signed.
In the \dmode{} mode Gemini-3.1 Pro scores 63.0\% on PostgreSQL and 61.4\% on Oracle,
while GPT-5.4 Mini scores 50.4\% and 52.9\%.
A5 Orchestration is the hardest subscenario for every model.
No model exceeds 49.8\% on it in the \dmode{} mode or 36.7\% in the \imode{} mode.
A5 asks the \solver{} to compose existing procedures into one transactional workflow
while respecting their signatures.

The two categories respond to interaction differently.
In the \dmode{} mode Synthesis and Editing are close:
59.8\% against 66.2\% for Gemini-3.1 Pro,
and 59.9\% against 64.4\% for GPT-5.5.
In the \imode{} mode they separate,
to 48.1\% against 73.5\% for Gemini-3.1 Pro.
Two subscenarios even score higher in the \imode{} mode.
A2 rises from 57.0\% to 68.1\% for Gemini-3.1 Pro
and from 49.8\% to 73.2\% for GLM-5.3,
and B3 rises from 67.5\% to 75.5\% for Gemini-3.1 Pro.

Two properties of the tasks account for this.
An Editing task hands the \solver{} the procedure it must change,
and that attachment already fixes most of what requirement perturbation removed,
so less has to be recovered by asking.
A2 and B3 are the two subscenarios whose difficulty is grounding rather than specification:
A2 must map a design document onto the real schema,
and B3 must reproduce a reported defect.
Both gain from the database tools the \imode{} mode provides,
and here that gain outweighs the missing requirement facts.
A Synthesis task offers no such anchor,
so it absorbs the full cost of an underspecified requirement.

\begin{figure*}[t]
\centering
\includegraphics[width=\textwidth]{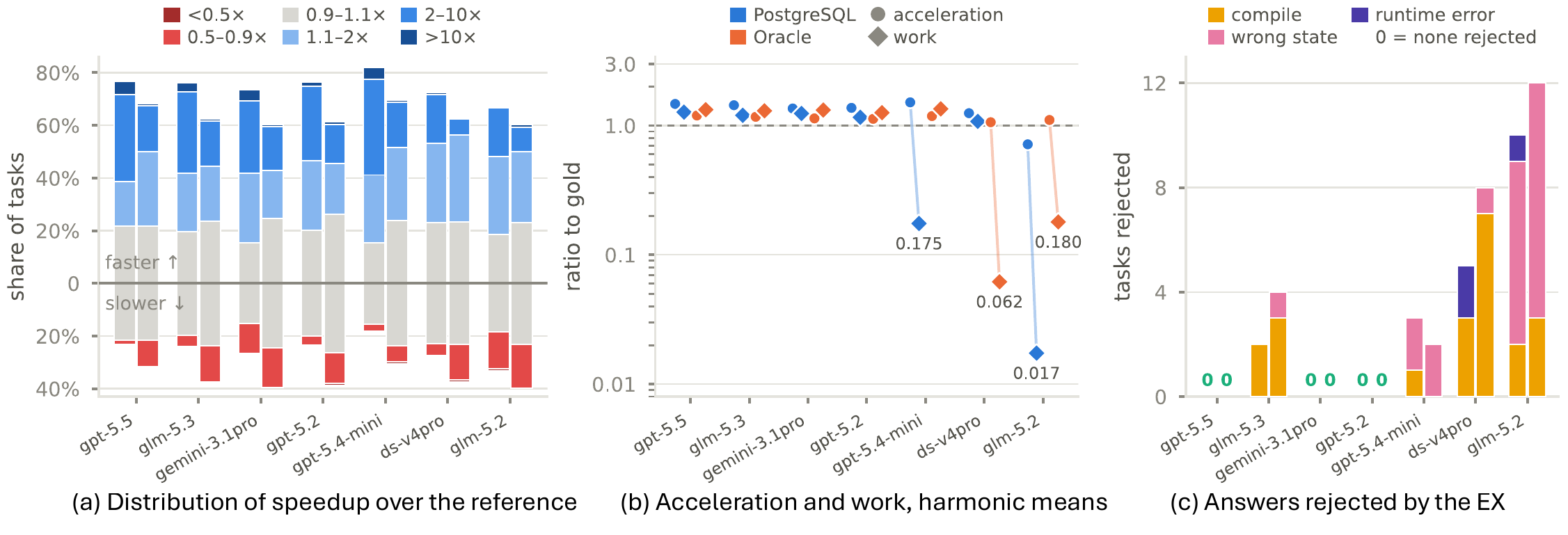}
\caption{\textbf{Most candidates match or beat the gold procedure, but two models buy their
speed with far more database work.}
Each model contributes one bar or marker pair per dialect, PostgreSQL then Oracle.
Panel~(a) stacks the faster speedup bins above the axis and the slower ones below.
Panel~(b) uses a log axis on which $1.0$ matches the gold procedure,
and its connector spans the acceleration and work means.
In panel~(c), a zero marks a model with no rejected task.}
\label{fig:b2}
\end{figure*}

\subsection{Performance Optimization}
\label{subsec:perf}

Figure~\ref{fig:b2} reports B2, the one subscenario that execution-state equivalence cannot rank;
the figure labels the gold procedure as the reference.
B2 hands the \solver{} a degraded version of the gold procedure
and asks for a faster one that behaves the same way.
The degradation applies real inefficiency patterns:
a set-based \texttt{UPDATE} becomes a cursor loop with one statement per row,
a join becomes an N+1 lookup,
and an aggregate is recomputed inside a loop.
The \solver{} is told that the procedure is correct but slow,
and is not told where the inefficiency is.
Seven models answer 120 tasks in each dialect, which gives 1{,}680 measurements.

Acceleration and work are defined in Section~\ref{subsec:setup};
what remains is how they are measured.
Every table is scaled to the row count recorded with the task,
inside a transaction that is rolled back afterwards.
Each procedure is then timed five times,
with the gold procedure, the degraded input and the candidate taking turns in rotating order,
so that cache warmth favors none of them.
The median of the five runs is reported.
We take the ratio between a procedure's own slowest and fastest run as the measurement noise,
whose median is $1.66$--$2.05$ on PostgreSQL and $1.42$--$1.48$ on Oracle.
Results are aggregated with the harmonic mean,
which penalizes a regression instead of letting one large win hide it.

Panel~(a) bins the tasks by speedup.
Every model matches or beats the gold procedure on most tasks in both dialects.
GPT-5.4 Mini places 36\% of its PostgreSQL tasks in the 2--10$\times$ band.
GLM-5.2 has the largest slower block, 14\% on PostgreSQL and 17\% on Oracle.
Panel~(b) reports the two harmonic means.
Five models sit just above the gold procedure on both.
GPT-5.4 Mini has the highest acceleration, $1.51$, but a work mean of $0.18$.
GLM-5.2 reaches $0.02$ on work,
and is the only model whose acceleration falls below the gold procedure, at $0.72$.
Panel~(c) counts the answers the semantic gate rejects,
split by compile failure, wrong final state, and runtime error.
GPT-5.5, Gemini-3.1 Pro and GPT-5.2 have none in either dialect.
GLM-5.2's rejections are mostly a wrong final state,
and DeepSeek V4 Pro's are mostly compile errors, seven of its eight on Oracle.
Only three of the 1{,}680 answers raised a runtime error.

Panel~(a) counts tasks and panel~(b) weighs them, which is why they disagree.
GLM-5.2's two worst PostgreSQL answers perform 4{,}445$\times$ and 1{,}058$\times$
the gold procedure's operations.
GPT-5.4 Mini's worst performs 516$\times$,
and its measured speedup is $0.97$,
close enough to the gold procedure to disappear into the matched band of panel~(a).
An answer can therefore look correct on the clock
while doing hundreds of times the database work.
On a scaled test table those extra operations cost little wall-clock time;
on production volumes they would not.
The same two models also account for most of the rejected answers,
and GLM-5.2's are changes of behavior rather than compile failures.
We read this as code that is locally plausible
without a model of what the database will actually do,
which is why B2 reports work beside time.

\subsection{Interaction Analysis}
\label{subsec:interaction}

\begin{figure}[t]
\centering
\includegraphics[width=\columnwidth]{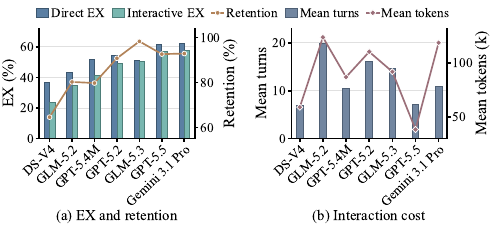}
\captionsetup{font=large}
\caption{\textbf{Interaction cost does not predict retention.}
Panel~(a): \dmode{} and \imode{} EX as bars, their ratio as a line.
Panel~(b): mean turns as bars, mean tokens per episode as a line.}
\label{fig:retention}
\end{figure}

\begin{figure}[t]
\centering
\includegraphics[width=\columnwidth]{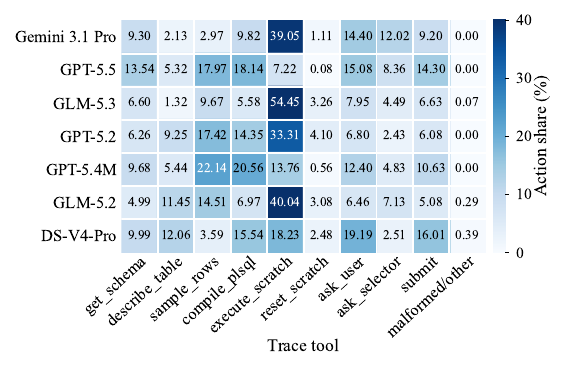}
\captionsetup{font=large}
\caption{\textbf{Models split between iterating against the database and inspecting it.}
Share of all recorded actions per model.
Tool names are verbatim from the traces;
names outside the protocol are grouped as \texttt{malformed/other}.}
\label{fig:action}
\end{figure}

\begin{figure}[t]
\centering
\includegraphics[width=\columnwidth]{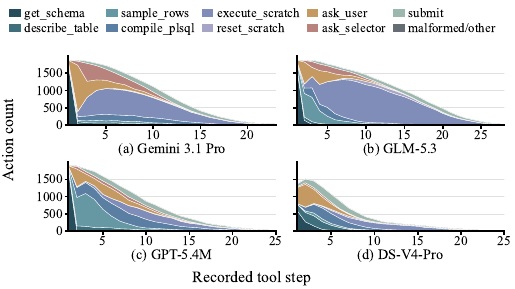}
\captionsetup{font=large}
\caption{\textbf{Models follow visibly different interaction trajectories.}
Action counts by recorded step, two models per row with shared axes.
The bands use the tool names of Figure~\ref{fig:action}.}
\label{fig:turns}
\end{figure}

This subsection asks how much accuracy survives requirement resolution,
what that costs, and what the cost is spent on.
Figure~\ref{fig:retention} plots \dmode{} and \imode{} EX as bars with their ratio as a line,
and mean turns as bars with mean total tokens as a line,
where tokens are the input plus output usage reported for an episode.
Figure~\ref{fig:action} gives each model's share of recorded actions over the nine tools,
plus one bucket for any action name outside the protocol.
Figure~\ref{fig:turns} gives action counts by recorded step
for four models spanning the \imode{} EX range.
It plots counts rather than shares,
so the outline of each stack is that model's episode attrition.
All three pool the paired PostgreSQL and Oracle tasks.

Retention is \imode{} EX divided by \dmode{} EX.
It measures how much of a model's own generation ability survives requirement resolution.
Raw \imode{} EX cannot measure this,
because a model with weak generation and strong elicitation
scores like a model with the opposite profile.

GLM-5.3 retains the most, 98.2\%, and DeepSeek V4 Pro the least, 64.5\%.
The other five fall between 79.4\% and 92.6\%.
Cost does not follow.
GPT-5.5 spends the fewest tokens, 38.5k per episode,
and the second fewest turns, 7.1, yet retains 92.6\%.
GLM-5.2 spends the most of both, 123.8k tokens over 19.8 turns, and retains 80.2\%.
DeepSeek V4 Pro spends the fewest turns, 6.9, and retains the least.

\begin{sloppypar}
Figure~\ref{fig:action} shows what the turns are spent on,
and the models split into two strategies.
GLM-5.3 spends 54.5\% of its actions on \texttt{execute\_scratch},
GLM-5.2 40.0\%, and Gemini-3.1 Pro 39.1\%:
these models iterate against a private copy of the database.
GPT-5.5 spends 7.2\% there,
and 18.0\% on \texttt{sample\_rows} and 18.1\% on \texttt{compile\_plsql} instead:
it reads the data and compiles the code rather than running it.
The two channels to the \usersimulator{} split the models the same way.
DeepSeek V4 Pro asks the most, 19.2\% of its actions on \texttt{ask\_user},
and confirms the least, 2.5\% on \texttt{ask\_selector}.
Gemini-3.1 Pro inverts this, at 12.0\% on \texttt{ask\_selector}, the highest in the figure.
Action names outside the protocol never exceed 0.4\%,
so tool-call formatting affects none of these numbers.
\end{sloppypar}

Figure~\ref{fig:turns} places the same behavior in time.
Gemini-3.1 Pro concentrates \texttt{ask\_user} in steps two to five and then stops asking.
GLM-5.3's \texttt{execute\_scratch} band runs to step 25,
and its episodes are the longest in the figure.
DeepSeek V4 Pro's episodes are nearly over by step ten.

Both strategies work.
GLM-5.3 iterates against the database and GPT-5.5 barely does,
yet they are the two models that retain the most, at 98.2\% and 92.6\%,
and GPT-5.5 reaches its figure on less than half the tokens.
Neither the volume of interaction nor the volume of questions predicts retention.
What fails is the absence of a strategy rather than the choice between them:
DeepSeek V4 Pro takes the fewest turns,
spends the largest share of them asking,
confirms the fewest readings,
and retains the least.
Retention is also unstable when the denominator is small,
so GLM-5.3's 98.2\% should be read against its \dmode{} EX of 51.3\%.

\subsection{Error Analysis}
\label{subsec:error}

\begin{figure}[t]
\centering
\includegraphics[width=0.95\columnwidth]{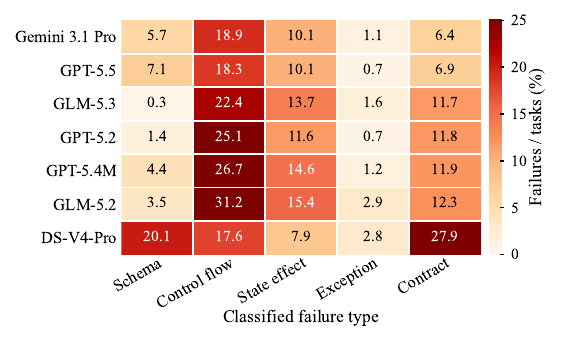}
\captionsetup{font=large}
\caption{\textbf{Control flow and persistent state dominate, and neither exists in
declarative SQL.}
Failures by error type in the \imode{} mode,
as a percentage of the tasks each model ran.}
\label{fig:error}
\end{figure}

Figure~\ref{fig:error} reports \imode{} failures by type,
as a percentage of the tasks each model ran,
so a cell reads directly against the \imode{} EX columns of Table~\ref{tab:main}.
Schema is a wrong table or column,
Control flow a wrong branch or loop,
State effect a wrong persistent write,
Exception a remaining runtime call failure,
and Contract an installation or interface mismatch.
We assign these types after evaluation with deterministic rules.
Schema and Contract come from installation, interface, argument, signature,
component-use, and database diagnostics,
and a remaining runtime call failure becomes Exception.
For a state mismatch,
a lexical comparison of branch, loop and return structure between the candidate and the
gold procedure separates Control flow from State effect.
This comparison is used only in post-hoc analysis
and is never shown to the \solver{}.

Control flow is the largest category for six of the seven models,
between 17.6\% and 31.2\%.
State effect is second, between 7.9\% and 15.4\%.
Exception is negligible everywhere, at most 2.9\%.
GLM-5.2 has the highest rate in both dominant categories, 31.2\% and 15.4\%.
Gemini-3.1 Pro and GPT-5.5 have the lowest rates in almost every category.
DeepSeek V4 Pro has a different profile:
its Schema rate is 20.1\% and its Contract rate is 27.9\%,
several times any other model's,
while its control-flow rate is the lowest at 17.6\%.

Models rarely crash.
They compile, they run, and they compute the wrong thing.
Two of the five categories have no analogue in declarative SQL,
because a query has no branch or loop to get wrong
and no persistent write to get wrong.
Together they account for between 25.5\% and 46.6\% of the tasks each model ran.
DeepSeek V4 Pro is the one model that fails before the logic rather than in it.
Its wrong tables and wrong signatures are consistent with its low retention
in Section~\ref{subsec:interaction},
and with the fact that it asks many questions but confirms few readings.

\subsection{Benchmark Validation}
\label{subsec:validation}
\label{subsec:realistic_ablation}

To verify the reliability of the \usersimulator{}, we conduct a human
audit of its \texttt{ask\_user} replies (Figure~\ref{fig:sim}). We sample
180 turns stratified over solver models and sub-scenarios, one third from
the LOC pool; auditors blind to model identity judged each reply against
the task's oracle evidence for acceptability, failure reason, and answer
leakage. Overall, 85.6\% of replies are acceptable and the residual
failures scatter over six categories with no dominant mode (a); LOC-pool
replies pass at a similar rate, with only 3.3\% leaking answer content (b).
The simulator thus provides reliable feedback with negligible answer
leakage.

\begin{table}[t]
\centering
\footnotesize
\setlength{\tabcolsep}{4pt}
\captionsetup{font=large}
\caption{\textbf{Every construction operation lowers EX on its own,
and the full pipeline lowers it most.}
Base is EX (\%);
indented rows are paired percentage-point changes against the same tasks.}
\label{tab:ablation}
\input{tables/ablation_body}
\end{table}

%% file: tables/direct_body.tex
DeepSeek V4 Pro & 39.5 & 36.1 & 27.1 & 44.5 & 33.3 & 40.3 & 19.2 & 41.7 & 35.2 & 46.1 & 34.5 & 41.7 & 44.8 & 30.5 & 44.1 & 15.0 & 50.9 & 38.3 \\
GLM-5.2 & 44.7 & 45.4 & 33.9 & 51.3 & 37.8 & 49.6 & 31.7 & 53.3 & 43.5 & 47.0 & 40.5 & 48.3 & 50.9 & 28.0 & 45.8 & 32.5 & 51.7 & 43.0 \\
GLM-5.3 & 62.3 & 52.1 & \underline{50.0} & 59.7 & 57.7 & \underline{61.3} & 29.2 & \underline{65.8} & 54.7 & 52.2 & \underline{47.4} & 52.5 & 49.1 & 28.0 & 52.5 & 46.7 & 54.3 & 47.8 \\
GPT-5.4 Mini & 57.9 & 52.9 & \underline{50.0} & 46.2 & 51.4 & 43.7 & 41.7 & 60.0 & 50.4 & \underline{63.5} & \textbf{55.2} & \underline{65.8} & 56.0 & 33.9 & 44.9 & 45.0 & 59.5 & 52.9 \\
GPT-5.2 & 64.9 & 52.9 & 44.1 & 60.5 & 58.6 & 47.1 & 38.3 & 65.0 & 53.8 & 62.6 & \textbf{55.2} & 63.3 & \underline{60.3} & \underline{34.7} & 52.5 & 49.2 & 62.9 & 55.1 \\
GPT-5.5 & \textbf{68.4} & \underline{58.0} & \textbf{55.1} & \underline{70.6} & \textbf{64.0} & 58.8 & \textbf{63.3} & \textbf{69.2} & \textbf{63.4} & \textbf{64.3} & \textbf{55.2} & \textbf{66.7} & \underline{60.3} & \textbf{36.4} & \underline{59.3} & \underline{68.3} & \textbf{67.2} & \underline{59.7} \\
Gemini-3.1 Pro & \underline{67.5} & \textbf{58.8} & \textbf{55.1} & \textbf{74.8} & \underline{59.5} & \textbf{64.7} & \underline{54.2} & \textbf{69.2} & \underline{63.0} & \textbf{64.3} & \textbf{55.2} & 64.2 & \textbf{62.1} & \textbf{36.4} & \textbf{62.7} & \textbf{80.8} & \underline{65.5} & \textbf{61.4} \\

%% file: tables/interactive_body.tex
DeepSeek V4 Pro & 5.3 & 10.9 & 5.1 & 10.1 & 12.6 & 28.6 & 29.2 & 33.3 & 17.0 & 29.6 & 52.6 & 28.3 & 35.3 & 19.5 & 29.7 & 7.5 & 42.2 & 30.5 \\
GLM-5.2 & 25.4 & 51.3 & 21.2 & 14.3 & 25.2 & 47.9 & 52.5 & 42.5 & 35.2 & 33.0 & 51.7 & 27.5 & 37.1 & 16.9 & 33.1 & 33.3 & 41.4 & 34.2 \\
GLM-5.3 & \textbf{53.5} & \textbf{73.9} & \underline{49.2} & 26.1 & \underline{31.5} & 60.5 & 55.0 & 53.3 & 50.5 & \underline{51.3} & 72.4 & 49.2 & 45.7 & 23.7 & 53.4 & 52.5 & 53.4 & 50.2 \\
GPT-5.4 Mini & 30.7 & 50.4 & 46.6 & 21.0 & 18.0 & 42.9 & 54.2 & 52.5 & 39.8 & 42.6 & 54.3 & 46.7 & 40.5 & 22.0 & 37.3 & 45.0 & 52.6 & 42.6 \\
GPT-5.2 & \underline{47.4} & \underline{68.1} & \textbf{55.1} & 31.1 & 30.6 & 58.0 & 60.8 & 60.8 & 51.7 & 41.7 & 69.0 & 46.7 & 49.1 & 22.0 & 42.4 & 44.2 & 62.1 & 47.1 \\
GPT-5.5 & 37.7 & 52.1 & 38.1 & \textbf{49.6} & \textbf{44.1} & \underline{71.4} & \underline{71.7} & \textbf{78.3} & \underline{55.6} & \textbf{57.4} & \textbf{79.3} & \underline{53.3} & \textbf{57.8} & \textbf{29.7} & \underline{54.2} & \underline{70.8} & \underline{64.7} & \underline{58.4} \\
Gemini-3.1 Pro & 39.5 & 61.3 & 31.4 & \underline{43.7} & \textbf{44.1} & \textbf{79.8} & \textbf{74.2} & \underline{75.0} & \textbf{56.4} & \underline{51.3} & \underline{75.0} & \textbf{54.2} & \underline{52.6} & \underline{28.0} & \textbf{63.6} & \textbf{76.7} & \textbf{71.6} & \textbf{59.1} \\

%% file: tables/ablation_body.tex
\begin{tabular*}{\columnwidth}{@{\extracolsep{\fill}}l cccc@{}}
\toprule
Condition & GPT-5.5 & Gemini-3.1 Pro & GPT-5.2 & GLM-5.2 \\
\midrule
Direct task & 86.67 & 82.92 & 72.92 & 73.33 \\
\quad + Knowledge injection & -16.27 & -9.09 & -13.40 & -25.36 \\
\quad + Contradiction injection & -10.92 & -13.87 & -10.92 & -15.13 \\
\quad + Requirement omission & -10.82 & -3.90 & -7.79 & -25.54 \\
Direct task + all three methods & -23.35 & -17.18 & -22.91 & -39.21 \\
\bottomrule
\end{tabular*}

%% file: contents/related_work.tex
\section{Related Work}
\label{sec:related}

\textbf{\nlplsql{}.}
To the best of our knowledge, PLForge \cite{zhang2025plforge} represents the first study specifically targeting \nlplsql{}, achieving SOTA performance through continued pre-training on a PL/SQL-centric corpus. However, PLForge relies on a template-based method for dataset generation, which may restrict syntactic diversity and limit its ability to reflect the complex distribution of real-world workloads. Although real-world procedural workloads also contain UDFs and triggers \cite{gupta2021procedural}, both PLForge and the current version of \benchmark{} focus on SPs. \benchmark{} instead broadens evaluation along a different dimension by covering multiple SP development scenarios in both direct and interactive settings.

\textbf{\nlsql{}.}
Recent progress on \nlsql{} is increasingly driven by language models, involving two main directions: supervised PLM-based encoder-decoder parsers and LLM-based approaches utilizing prompting and workflows.
\textit{PLM-based encoder-decoder parsers} typically map serialized inputs to SQL. Key enhancements include injecting graph structures for schema-aware learning \cite{Cao2021LGESQLLG,Li2023GraphixT5MP}, utilizing intermediate representations to decouple schema linking \cite{Guo2019TowardsCT,Gan2021NaturalSM,Li2023RESDSQLDS}, and applying constrained decoding to ensure syntactic validity \cite{Scholak2021PICARDPI}. These methods establish reliable baselines for encoding and constraining schema information. \textit{LLM-based \nlsql{}} methods have shifted \nlsql{} toward prompt-centric and agentic paradigms \cite{Luo2025NaturalLT,Hong2024NextGenerationDI,Liu2024ASO}, including in-context learning (ICL), fine-tuning (FT), and agentic workflows. ICL exploits prompting via decomposition and execution-guided refinement \cite{Liu2024ASO,Hong2024NextGenerationDI,pourreza2024din}, improving schema grounding through retrieval and dynamic few-shot construction \cite{Li2025AIDSQLAI,Xie2025OpenSearchSQLET,Ren2024PURPLEMA}. FT approaches adapt models for controllability, leveraging domain-oriented training \cite{Li2024CodeSTB,sterbentz2026ringsql,lhy2025omnisql,park2026resql} and structural constraints \cite{zhang2025structureguided,Ren2025ThePO}. Agentic workflows treat synthesis as an iterative process involving planning and feedback, exemplified by MCTS-based Alpha-SQL \cite{li2025alphasql}, multi-agent collaboration \cite{Wang2023MACSQLAM}, and user-in-the-loop clarification \cite{Zhao2024SphinteractRA,mazumder2026atomsql}. Although BIRD-INTERACT \cite{huo2026bird} introduces dynamic and stateful multi-turn evaluation for Text-to-SQL, it remains focused on SQL statements and does not evaluate dialogue-based alignment and revision of procedural PL/SQL specifications.

\textbf{Code Language Models.}
Existing code language models \cite{lozhkov2024starcoder,roziere2023code,hui2024qwen2,jiang2024survey} rely on massive polyglot training data to achieve broad coverage. However, this general-purpose objective often leads to suboptimal performance on specialized, low-resource languages such as PL/SQL. The scarcity of high-quality PL/SQL data in public repositories remains a major obstacle to PL/SQL-specific model adaptation. Rather than proposing a new model-specialization method, \benchmark{} complements existing model-centric research with a model-agnostic, executable benchmark for evaluating procedural database programming across multiple development and interaction settings.

%% file: contents/conclusion.tex
\begin{figure}[t]
\centering
\includegraphics[width=\columnwidth]{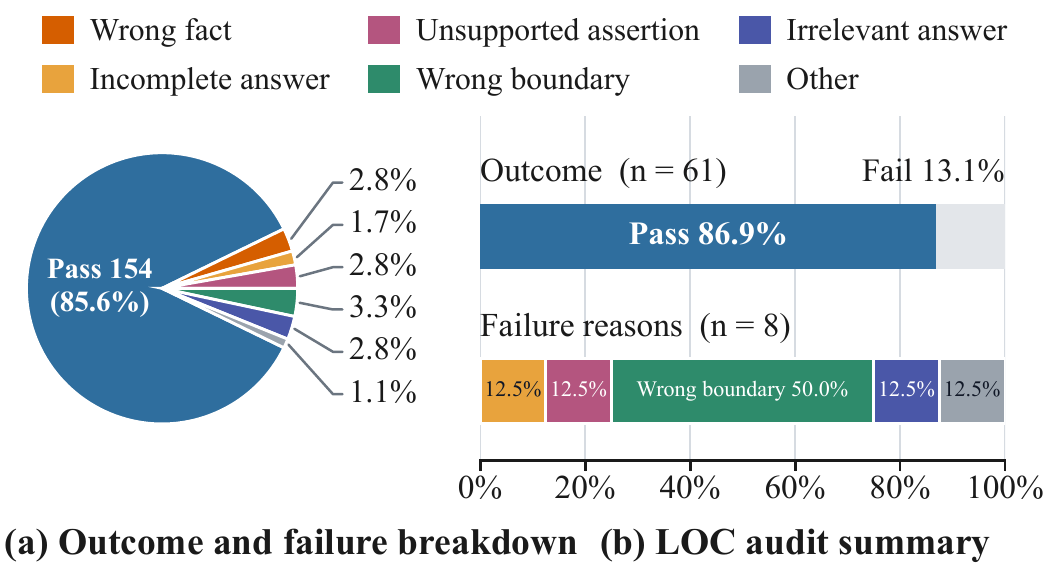}
\captionsetup{font=large}
\caption{\textbf{Human audit}
(a) Pass rate and failure reasons over all audited \texttt{ask\_user} replies.
(b) The same audit restricted to the sampled LOC pool.}
\label{fig:sim}
\end{figure}

\section{Conclusion}
\label{sec:conclusion}

We proposed \benchmark{}, to our knowledge the first benchmark to evaluate LLMs on NL-to-PL/SQL across multiple development scenarios in both \dmode{} and \imode{} modes:
an executable suite of 3{,}998 tasks spanning nine \dmode{} subscenarios and eight paired \imode{} subscenarios in two dialects.
Iterative Logic Enhancement, live validation, and scenario-specific adapters constructed verifiable \dmode{} tasks,
from which Knowledge Integration and Requirement Perturbation derived paired \imode{} tasks whose executable target an isolated Offline Evaluator preserved.
Across seven models, the strongest reached 62.2\% execution accuracy in the \dmode{} mode and 57.8\% in the \imode{} mode,
interaction cost accuracy on Synthesis and recovered it on Editing,
and failures concentrated in control flow and persistent state, the two error types that declarative SQL does not have.
We released the benchmark, the interaction traces, and the construction pipeline for reproducible evaluation.